\documentclass[runningheads]{llncs}
\usepackage[T1]{fontenc}
\usepackage{graphicx}
\usepackage{pifont}
\usepackage{caption}
\usepackage{subcaption}
\usepackage{color}
\usepackage{enumitem}
\newcommand{\cmark}{\ding{51}}%
\newcommand{\xmark}{\ding{55}}%
\begin{document}
\title{KeepOriginalAugment: Single Image-based Better Information-Preserving Data Augmentation Approach}
\titlerunning{KeepOriginalAugment}
%
\author{Teerath Kumar\inst{1} \and
Alessandra Mileo\inst{2} \and
Malika Bendechache\inst{3}}
\authorrunning{T. Kumar et al.}
%
\institute{CRT-AI \& ADAPT Research Centre,
School of Computing,
Dublin City University, Dublin, Ireland \and
INSIGHT \& I-Form Research Centre,
School of Computing,
Dublin City University, Dublin , Ireland
\and
ADAPT \& Lero Research Centres,
School of Computer Science,
University of Galway, Galway, Ireland}
\maketitle              
\begin{abstract}
Advanced image data augmentation techniques play a pivotal role in enhancing the training of models for diverse computer vision tasks. Notably, SalfMix and KeepAugment have emerged as popular strategies, showcasing their efficacy in boosting model performance. However, SalfMix reliance on duplicating salient features poses a risk of overfitting, potentially compromising the model's generalization capabilities. Conversely, KeepAugment, which selectively preserves salient regions and augments non-salient ones, introduces a domain shift that hinders the exchange of crucial contextual information, impeding overall model understanding. In response to these challenges, we introduce KeepOriginalAugment, a novel data augmentation approach. This method intelligently incorporates the most salient region within the non-salient area, allowing augmentation to be applied to either region. Striking a balance between data diversity and information preservation, KeepOriginalAugment enables models to leverage both diverse salient and non-salient regions, leading to enhanced performance. We explore three strategies for determining the placement of the salient region—minimum, maximum, or random — and investigate swapping perspective strategies to decide which part (salient or non-salient) undergoes augmentation. Our experimental evaluations, conducted on classification datasets such as CIFAR-10, CIFAR-100, and TinyImageNet, demonstrate the superior performance of KeepOriginalAugment compared to existing state-of-the-art techniques. The source code for our KeepOriginalAugment method, along with trained models, is publicly available at \url{https://github.com/kmr2017}.

\end{abstract}
\section{Introduction}
Data augmentation has proven to be an essential technique for enhancing model generalization in various challenging computer vision tasks, including image classification \cite{simonyan2014very,he2016deep,szegedy2015going,szegedy2016rethinking}, image segmentation \cite{long2015fully,zhao2017pyramid}, and object detection \cite{girshick2014rich,girshick2015fast,ren2015faster}. The diversity of data provided by data augmentation methods plays a crucial role in enabling models to consistently perform well on both training and test data. In recent years, several label-invariant data augmentation methods have been proposed to further enhance the effectiveness of data augmentation \cite{devries2017improved,zhong2020random,chen2020gridmask,kumar2017hide}. These methods include techniques such as cutout, random erasing, grid mask, and hide-and-seek. Additionally, reinforcement learning-based approaches, such as autoaugment \cite{cubuk2019autoaugment}, have been developed to discover optimal data augmentation policies. Non-reinforcement learning-based methods, such as SalfMix \cite{choi2021salfmix}, KeepAugment \cite{gong2021keepaugment}, and Randaugment \cite{cubuk2020randaugment}, have also addressed computational and feature fidelity challenges. However, it is important to note that these methods, while promoting data diversity, can introduce noise and reduce feature fidelity, thus impacting overall model performance. For instance, SalfMix can lead to overfitting by repeating the salient part in the image \cite{choi2021salfmix}, while KeepAugment can introduce a domain shift between salient and non-salient regions, hindering contextual information \cite{gong2021keepaugment}.
To address these challenges and simultaneously increase feature fidelity and diversity, we propose a simple yet effective data augmentation technique named KeepOriginalAugment. Our approach detects the salient region and incorporates it into a non-salient region, allowing augmentation on either or both regions. Note, the model and neural network are used interchangeably in the rest of  work. 

The rest of the paper is organized as follows: Section \ref{motivation} describes motivation, Section \ref{related_work} provides a review of related literature, Section \ref{proposed_method} presents the details of the proposed KeepOriginalAugment method, Section \ref{experiments} describes the experimental setup and presents the results, and finally, Section \ref{conclusion} concludes the paper.

\begin{figure*}[htbp]

        \includegraphics[page=1, width=1.2\textwidth, height=3.5cm, bb=100bp 150bp 1000bp 400bp, clip]{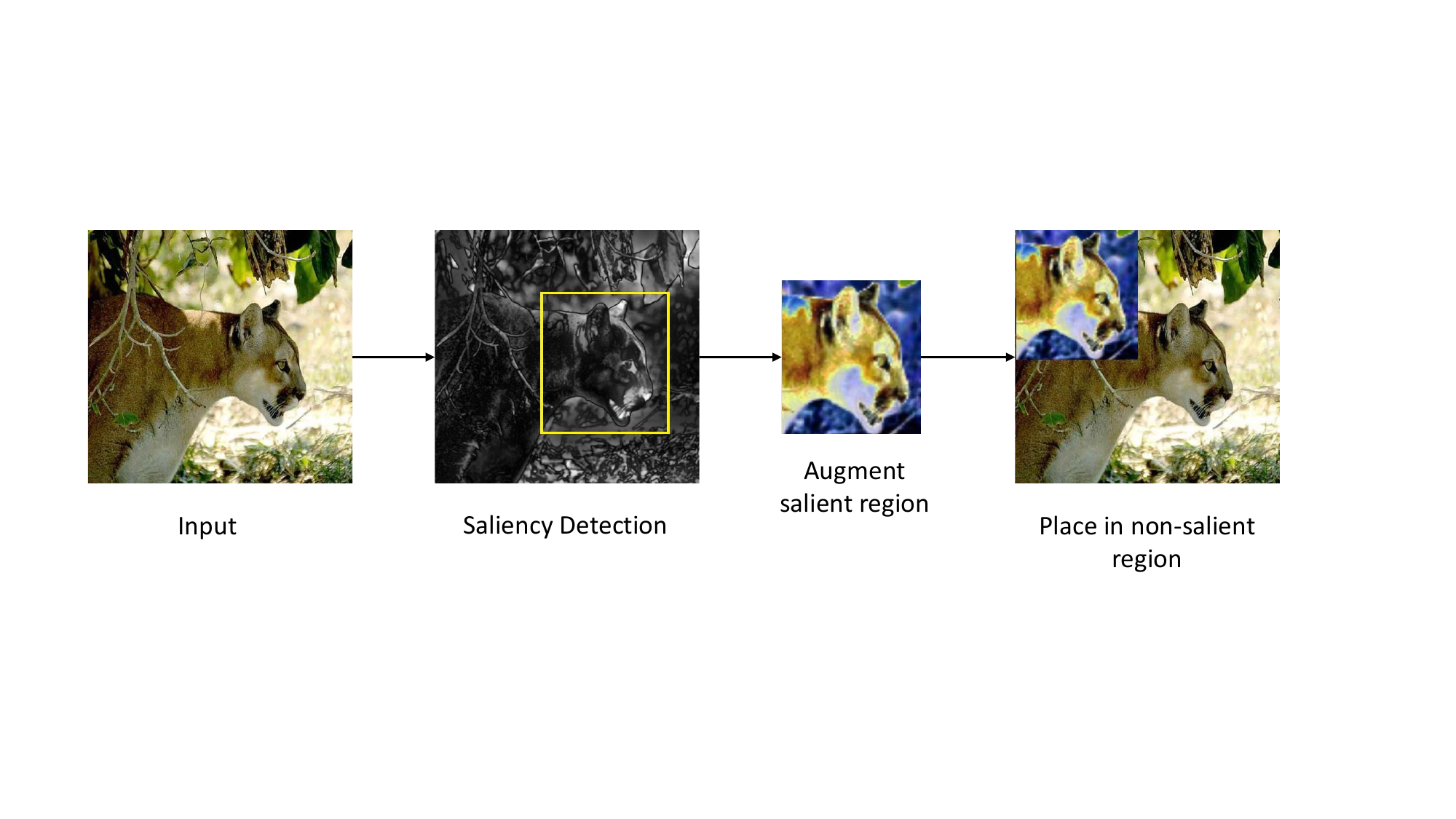}
    \caption{Overall KeepOriginalAugment process}
    \label{fig:KeepOriginalAug}
\end{figure*}
{

\begin{figure*}[!ht]
    \centering
    \begin{subfigure}{0.22\textwidth}
        \includegraphics[width=\textwidth, height=3cm]{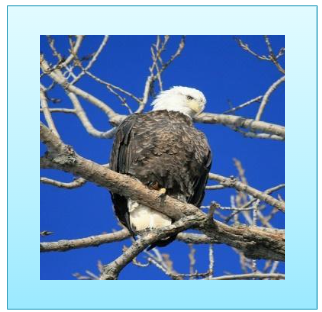}
        \caption{Input}
        \label{fig:Input}
    \end{subfigure}%
    \hfill
    \begin{subfigure}{0.22\textwidth}
        \includegraphics[width=\textwidth, height=3cm]{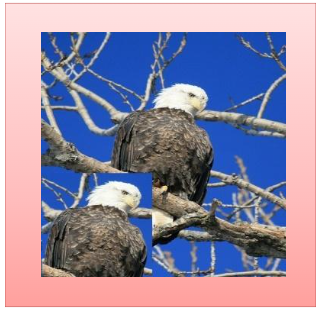}
        \caption{SalfMix}
        \label{fig:Salfmix}
    \end{subfigure}%
    \hfill
    \begin{subfigure}{0.22\textwidth}
        \includegraphics[width=\textwidth, height=3cm]{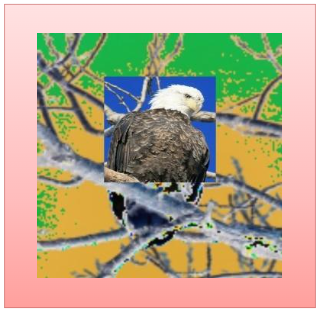}
        \caption{KeepAugment}
        \label{fig:KeepAugment}
    \end{subfigure}%
    \hfill
    \begin{subfigure}{0.22\textwidth}
        \includegraphics[width=\textwidth, height=3cm]{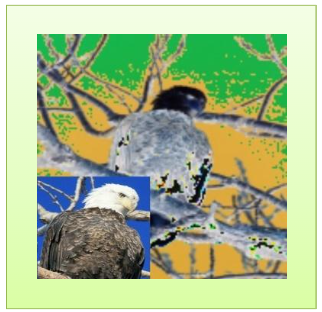}
        \caption{Ours}
        \label{fig:ours}
    \end{subfigure}
    \captionsetup{belowskip=-10pt}
    \caption{Comparison of the relevant data augmentation methods with ours}
    \label{rfidtag_testing}
\end{figure*}

  

\begin{table*}

  \caption{Comparison of the proposed KeepOriginalAugment with existing SOTA augmentation, where Mixed represents images are mixed or not, one-image represents method uses one image, Saliency represents a particular method uses saliency, salient region augmented (SRA) only salient part is augmented or not, Non-salient region augmented (NSRA) means only non-salient regions are augmented and blending represents images are blended.     }
  \label{tab:comparison}
  \begin{tabular}{|c|c|c|c|c|c|c|l|}
  \hline 
  \textbf{Method} & \textbf{Mixed} ? & \textbf{One-Image?} & \textbf{Saliency?} & \textbf{SRA?} & \textbf{NSRA?}  & \textbf{Blending?} \\ \hline 
 
Mixup~\cite{zhang2017mixup} & \textcolor{red}{\xmark} & \textcolor{red}{\xmark} & \textcolor{red}{\xmark} & \textcolor{red}{\xmark} & \textcolor{red}{\xmark} &  \textcolor{green}{\cmark} \\ \hline 

CutMix~\cite{yun2019cutmix} & \textcolor{green}{\cmark} & \textcolor{red}{\xmark} & \textcolor{red}{\xmark} & \textcolor{red}{\xmark} & \textcolor{red}{\xmark} &  \textcolor{red}{\xmark} \\  \hline 
SaliencyMix~\cite{uddinsaliencymix} & \textcolor{green}{\cmark} & \textcolor{red}{\xmark} & \textcolor{green}{\cmark} & \textcolor{red}{\xmark} & \textcolor{red}{\xmark} &  \textcolor{red}{\xmark} \\ \hline 
ResizeMix~\cite{qin2020resizemix} & \textcolor{green}{\cmark} & \textcolor{red}{\xmark} & \textcolor{red}{\xmark} & \textcolor{red}{\xmark} & \textcolor{red}{\xmark}  & \textcolor{red}{\xmark} \\ \hline 
PuzzleMix~\cite{kim2020puzzle} & \textcolor{green}{\cmark} & \textcolor{red}{\xmark} & \textcolor{green}{\cmark} & \textcolor{red}{\xmark} & \textcolor{red}{\xmark}   & \textcolor{green}{\cmark}  \\ \hline 
Self-Augmentation~\cite{seo2021self} & \textcolor{green}{\cmark} & \textcolor{green}{\cmark} & \textcolor{red}{\xmark} & \textcolor{red}{\xmark} & \textcolor{red}{\xmark} &  \textcolor{red}{\xmark} \\ \hline 
SalfMix~\cite{choi2021salfmix} & \textcolor{green}{\cmark} & \textcolor{green}{\cmark} & \textcolor{green}{\cmark} & \textcolor{red}{\xmark} & \textcolor{red}{\xmark} &  \textcolor{red}{\xmark} \\ \hline 
KeepAugment~\cite{gong2021keepaugment} & \textcolor{green}{\cmark} & \textcolor{green}{\cmark} & \textcolor{green}{\cmark} & \textcolor{red}{\xmark} & \textcolor{green}{\cmark} &  \textcolor{red}{\xmark} \\ \hline 
Ours& \textcolor{green}{\cmark} & \textcolor{green}{\cmark} & \textcolor{green}{\cmark} & \textcolor{green}{\cmark} & \textcolor{green}{\cmark} &  \textcolor{red}{\xmark} \\ \hline

  
\end{tabular}

\end{table*}

\section{Motivation}\label{motivation}
The field of computer vision has witnessed remarkable advancements in recent years, driven in large part by innovative data augmentation techniques aimed at enhancing model generalization. These techniques, such as SalfMix and KeepAugment, have shown their efficacy in improving model performance. However, they are not without their limitations. SalfMix, a technique that duplicates salient features, can inadvertently lead to overfitting~\ref{fig:Salfmix}, diminishing a model's ability to generalize beyond the training data. Similarly, KeepAugment, which selectively preserves salient regions during data augmentation, can introduce a domain shift between salient and non-salient regions~\ref{fig:KeepAugment}, obstructing the seamless exchange of vital contextual information and hampering the overall model's understanding.

The need for a more balanced approach that harnesses both diverse salient and diverse non-salient regions for augmentation becomes evident. This balance is crucial to empower models to achieve higher performance improvements without compromising on information preservation.

In light of these challenges, our research introduces a novel data augmentation approach: KeepOriginalAugment. The motivation behind KeepOriginalAugment lies in its ability to strike an optimal balance between data diversity and information preservation. By intelligently incorporating the most salient region within the non-salient region, it enables augmentation to be applied to either or both regions. This approach holds the promise of significantly enhancing model performance, as our experiments on benchmark datasets demonstrate. In a landscape where data augmentation is pivotal to model success, our research seeks to address the limitations of existing techniques, contributing a novel solution that empowers models to leverage the full potential of diverse data while maintaining feature fidelity. 

In summary, the contributions of our work are as follows:
\begin{itemize}
\item We propose KeepOriginalAugment as a solution to the limitations of SalfMix and KeepAugment.
\item We validate the effectiveness of our approach through experiments on various datasets using different network architectures.
\item We provide the code on github.
\end{itemize}
\section{Related Work}\label{related_work}
Several data augmentation methods~\cite{kumar2021binary,kumar2023rsmda,turab2022investigating,chandio2021audd,kumar2023image,kumar2021class} have been proposed to enhance model performance in computer vision tasks. In this paper, we focus on closely related data augmentation techniques that have been successful in achieving state-of-the-art (SOTA) results. More precisely, KeepOriginalAugment has been differentiated from existing SOTA methods in table~\ref{tab:comparison}. Furthermore, for the sake of easiness, we classify the data augmentation into two subcategories. 
\subsection{Information Erasing Augmentation}
Information erasing augmentation focuses on removing specific information from images to induce occlusion and compel models to learn the erased features. Various methods have been proposed in this category, including random erasing \cite{zhong2020random}, cutout \cite{devries2017improved}, grid mask \cite{chen2020gridmask}, and hide-and-seek \cite{kumar2017hide}. Cutout randomly masks out regions in the image, while random erasing dynamically determines the regions to be masked based on aspect ratio and size. Grid mask applies a uniform masking pattern to erase information, while hide-and-seek divides the image into uniform squares and randomly masks out a varying number of squares.
Although these methods force deep learning models to learn the erased features, they can introduce challenges. On one hand, there is a risk of generating noisy examples that can negatively impact performance. On the other hand, removing contextual information from the image can lead to overfitting. This loss of feature fidelity and diversity hampers the overall effectiveness of these methods as shown in Figure \ref{fig:noisy_data} for an example of noisy data and Figure \ref{fig:overfitting} for an example of overfitting. To address these limitations, it is crucial to explore information-preserving augmentation techniques that promote diversity without compromising feature fidelity.
\begin{figure*}[htbp]
    \centering
    \begin{subfigure}{0.22\textwidth}
        \includegraphics[width=\textwidth, height=3cm]{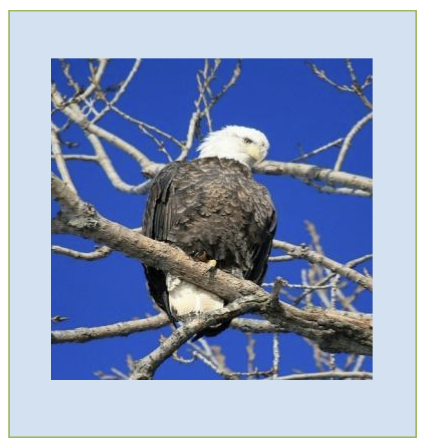}
        \caption{Input}
        \label{fig:Input1}
    \end{subfigure}
    \hfill
    \begin{subfigure}{0.22\textwidth}
        \includegraphics[width=\textwidth, height=3cm]{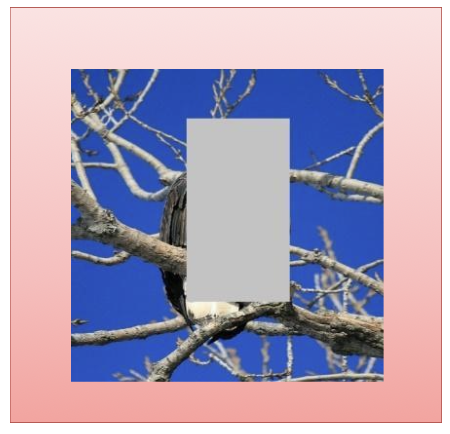}
        \caption{RE}
        \label{fig:RE_noisy}
    \end{subfigure}
    \hfill
    \begin{subfigure}{0.22\textwidth}
        \includegraphics[width=\textwidth, height=3cm]{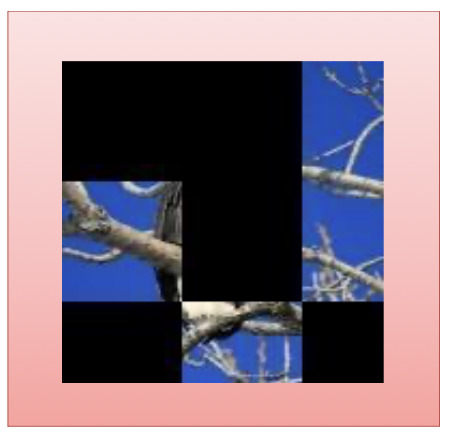}
        \caption{Hide-and-Seek}
        \label{fig:HaS_noisy}
    \end{subfigure}
    \hfill
     \begin{subfigure}{0.22\textwidth}
        \includegraphics[width=\textwidth, height=3cm]{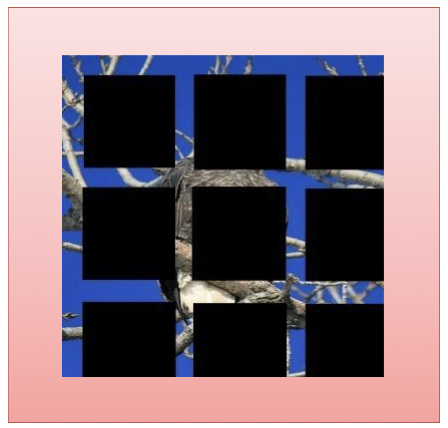}
        \caption{GridMask}
        \label{fig:GM_noisy}
    \end{subfigure}
     \captionsetup{belowskip=-10pt}
    \caption[]{Are these augmented examples helpful for training?}
    \label{fig:noisy_data}
\end{figure*}
\begin{figure*}[htbp]
    \centering
    \begin{subfigure}{0.22\textwidth}
        \includegraphics[width=\textwidth, height=3cm]{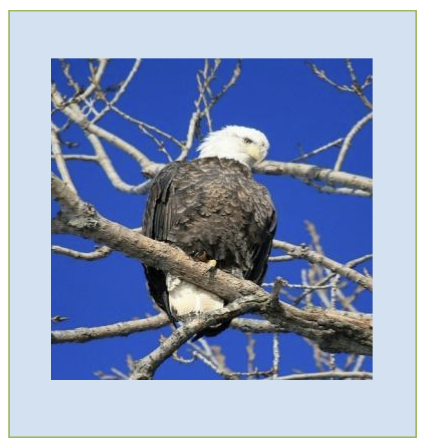}
        \caption{Input}
        \label{fig:Input2}
    \end{subfigure}
    \hfill
    \begin{subfigure}{0.22\textwidth}
        \includegraphics[width=\textwidth, height=3cm]{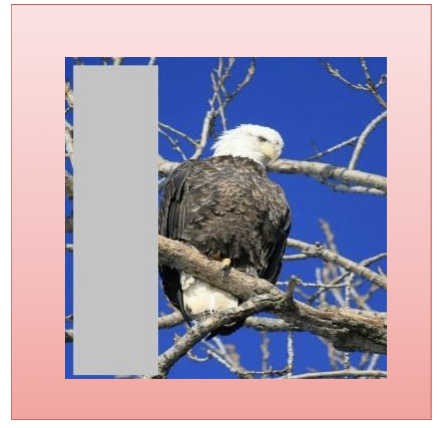}
        \caption{RE}
        \label{fig:RE_overfitting}
    \end{subfigure}
    \hfill
    \begin{subfigure}{0.22\textwidth}
        \includegraphics[width=\textwidth, height=3cm]{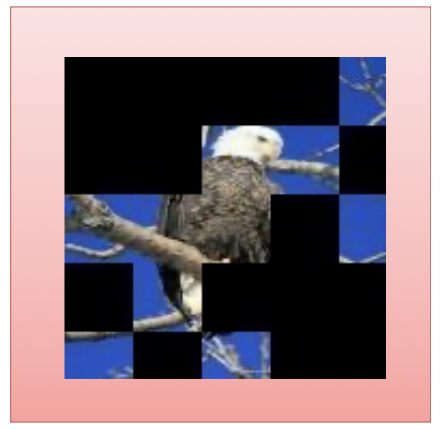}
        \caption{Hide-and-Seek}
        \label{fig:HaS_overfitting}
    \end{subfigure}
    \hfill
     \begin{subfigure}{0.22\textwidth}
        \includegraphics[width=\textwidth, height=3cm]{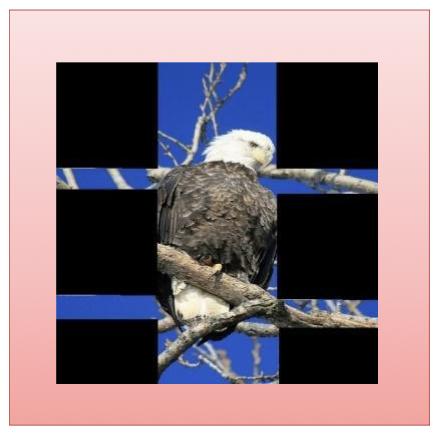}
        \caption{GridMask}
        \label{fig:GM_overfitting}
    \end{subfigure}
     \captionsetup{belowskip=-10pt}
    \caption[]{Are these augmented examples helpful for model generalization?}
    \label{fig:overfitting}
\end{figure*}
    \begin{figure*}[htbp]
        \includegraphics[page=2, width=1.0\textwidth, height=6.5cm, bb=0bp 50bp 900bp 500bp, clip]{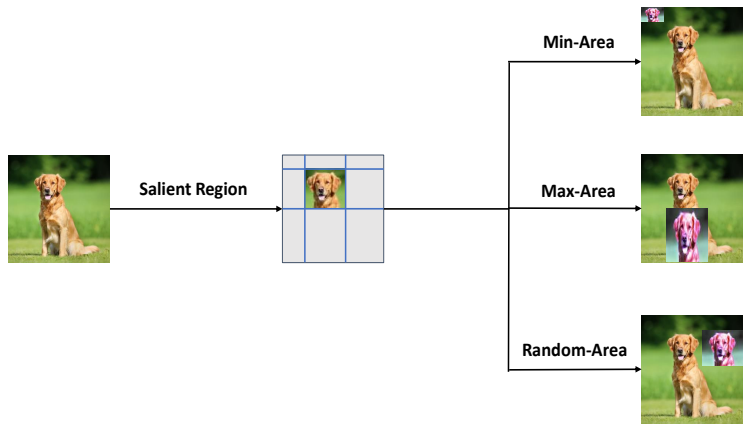}
         \captionsetup{belowskip=-10pt}
    \caption{Strategies: where to place the salient region?}
    
    \label{fig:where_to_place}
\end{figure*}
\subsection{Information Preserving Augmentation}
Information-preserving augmentation methods are designed to promote diversity during the training process while preserving important information. Two notable techniques in this category are AutoAugment and RandAugment. AutoAugment employs reinforcement learning and policy gradient optimization to automatically search for effective data augmentation policies. On the other hand, RandAugment reduces the computational complexity of AutoAugment by randomly selecting data augmentation policies, yet it still achieves impressive results without the need for reinforcement learning \cite{cubuk2020randaugment}.
In addition to these methods, there are closely related data augmentations called SalfMix~\cite{choi2021salfmix} and KeepAugment \cite{gong2021keepaugment}. SalfMix detects the salient region in an image and transfers it to a non-salient region, forcing the model to learn the salient features. However, this approach may lead to overfitting due to the redundancy of the same salient feature, as illustrated in Figure \ref{fig:Salfmix}. KeepAugment, on the other hand, detects the salient region and preserves it without any augmentation, while applying RandAugment only to the non-salient region. This approach promotes feature fidelity by preserving important information, but it introduces a domain shift problem between the salient and non-salient regions. Consequently, the contextual information between these regions is hindered, as depicted in Figure \ref{fig:KeepAugment}. Our approach differs from KeepAugment in two key aspects. Firstly, in KeepAugment, non-salient regions remain unchanged during data augmentation, whereas our approach allows augmentation of these regions. Secondly, KeepAugment fixes the placement of salient regions, whereas our proposed method varies the placement to achieve scaled augmentation.

    \begin{figure*}[hbtp]
        \includegraphics[page=3, width=0.98\textwidth, height=6.5cm, bb=0bp 50bp 900bp 500bp, clip]{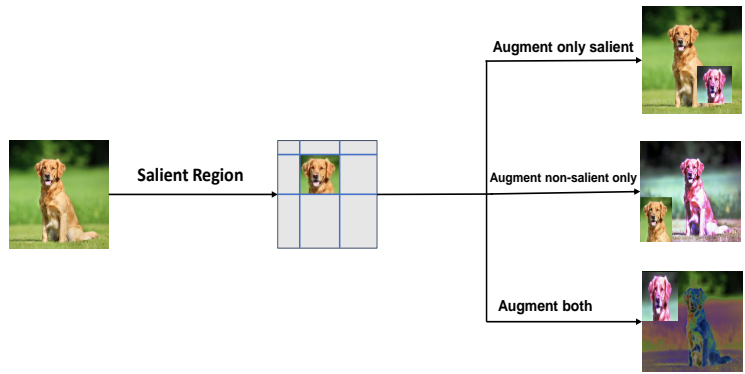}
    \caption{Strategies: Which part should be augmented?}
    \label{fig:which_part_to_augment}
\end{figure*}
\section{Proposed method}\label{proposed_method}

Our primary objective is to enhance fidelity by preserving both the original and augmented information within a single image, thereby encouraging the model to learn diverse information. To achieve this, we begin by measuring the importance of different regions in the image using a saliency map. Let $s_\mathrm{(i,j)}(x,y)$ represent the saliency map of image $x$ with label $y$ at pixel $(i,j)$. We calculate the importance score $I$ for a specific region $R$ as follows:

\begin{equation}
I(R,x,y) = \sum_{(i,j) \in R} s_\mathrm{(i,j)}(x,y)
\label{eq:importance_equ}
\end{equation}

In this equation, we sum up the saliency scores of all pixels within the region $R$. We employ a standard saliency map based on the vanilla gradient method proposed by Simonyan et al. \cite{simonyan2014deep}. Next, we preserve only those regions that possess an importance score $I$ greater than a predefined threshold $\tau$. We used same $\tau$ as it is used in KeepAugment~\cite{gong2021keepaugment}. By identifying these important regions, we aim to address the issue of redundant salient features present in SalfMix, as well as tackle the domain shift problem encountered in KeepAugment:
\begin{enumerate}[label=(\alph*)]
\item \textbf{Where to place the salient region?}
To determine the placement of the salient region within the non-salient region and achieve diversity, we explore three different strategies:

\textbf{Min-Area:} In this strategy, we identify eight regions surrounding the salient region. Among these regions, we select the one with the minimum area. We then resize the salient region according to the size of that minimum area and place it within that region. This is illustrated in Figure~\ref{fig:where_to_place}

\textbf{Max-Area:} Conversely, in the Max-Area strategy, we identify the eight regions and choose the one with the maximum area. The salient region is resized accordingly and placed within this region. This is demonstrated in Figure~\ref{fig:where_to_place}.

\textbf{Random-Area:} To introduce scaling augmentation to the salient region, we adopt a more flexible approach. Instead of limiting the placement to the minimum or maximum area, we randomly select one of the eight regions. The salient region is resized based on the size of the chosen region, and it is placed within that area. This is depicted in Figure~\ref{fig:where_to_place}.

Through experimental evaluations, we have observed that the Random-Area strategy provides greater diversity. This is attributed to the random scaling data augmentation applied to the salient region, which introduces varying degrees of augmentation effects.


  \item \textbf{Which part should be augmented?}
After identifying the salient region, we propose and investigate three distinct strategies to enhance fidelity and diversity:

\textbf{Augment only salient:} In this strategy, we solely apply random augmentation to the salient region. Subsequently, we paste the augmented salient region into the non-salient region of the original image. It is crucial to emphasize that augmentation is exclusively performed on the salient region. The application of this strategy is illustrated in Figure~\ref{fig:which_part_to_augment}.

\textbf{Augment non-salient only:} In this strategy, we conduct augmentation on the entire image while preserving the original salient region. The augmented image is then combined with the original salient region, which is extracted from the unaltered original image. The overall approach is depicted in Figure~\ref{fig:which_part_to_augment}.

\textbf{Augment both:} This strategy involves performing separate augmentations on both the salient region and the entire image. The augmented salient region is integrated with the augmented whole image, as shown in Figure~\ref{fig:which_part_to_augment}.
By considering a trade-off probability between the original sample and the augment both strategy sample, we observed that the augment both strategy demonstrates greater diversity and fidelity across various computer vision tasks as discussed in section~\ref{experiments}. It is important to note that we utilize randAug~\cite{cubuk2020randaugment} for augmentation, which offers computational efficiency similar to that employed by KeepAugment. 
   
\end{enumerate}

\section{Experiments}\label{experiments}

\subsection{Training setup}

For a fair comparison, we followed the training setups outlined in previous works, specifically KeepAugment~\cite{gong2021keepaugment}, SalfMix~\cite{choi2021salfmix}, and random erasing~\cite{zhong2020random}. Our training process consisted of 300 epochs, a batch size of 128, an initial learning rate of 0.1, and the CosineAnnealingLR scheduler. We employed stochastic gradient descent with a momentum of 0.9 and weight decay of 0.0005. To evaluate the generalization performance, we utilized various neural network architectures such as ResNet, Wide-ResNet, and PreActResNet.

To assess the effectiveness of our proposed methods, we conducted experiments on different datasets, including CIFAR-10 (10 classes), CIFAR-100 (100 classes)~\cite{krizhevsky2009learning}, and TinyImageNet (200 classes) ~\cite{le2015tiny}.

For the PreActResNet architecture, we employed the standard saliency function from the OpenCV library\footnote{$https://docs.opencv.org/3.4/da/dd0/classcv\_1\_1saliency\_1\_1StaticSaliency \\ FineGrained.html$} since PreActResNet required more computation time for saliency estimation.

Throughout our experiments, we utilized accuracy and error rate as performance metrics. Higher accuracy values indicate better performance, while lower error rates are desired. For dataset diversity measurement and query retrieval results, we followed the same parameters as discussed in~\cite{mandal2021dataset}.
\subsection{Hyperparamester}

To determine the optimal combination of salient region placement strategy and region augment strategy, we conducted experiments using the ResNet-18 neural network architecture on the CIFAR-10 dataset. 

Based on our experiments, we found that the augment both strategy, combined with the Random-Area strategy, yielded the best results in terms of performance. The reasons behind the effectiveness of this combination are discussed in Section~\ref{discussion}.
\subsection{Results}

In Table~\ref{tab:cifar10_results}, we present the performance of our methods, including KeepOriginalCutout, compared to the baseline and other state-of-the-art techniques, such as Cutout and KeepCutout, on the CIFAR-10 dataset. It is important to mention KeepOriginalCutout is combination of the cutout and KeepOriginalAugment, which basically cutout the non-salient region then combine with KeepOriginalAugment like KeepAugment~\cite{gong2021keepaugment}. Our method, KeepOriginalCutout, demonstrated superior accuracy, with an absolute 1\% improvement over the baseline and a 0.6\% improvement over the KeepCutout method, when evaluated using the ResNet-18 architecture. Additionally, our method showed competitive accuracy performance when evaluated using the ResNet-110 and Wide ResNet architectures.

To assess the generalization of our proposed approach, we conducted experiments on larger and more diverse datasets using various neural network architectures. The results, presented in Table~\ref{tab:all_datasets_results}, show that our proposed method achieved superior error rate performance compared to all other data augmentation methods. It even outperformed different versions of HybridMix, which combines several image-mixing methods as an ensemble, in most cases. However, there were a few instances where our proposed method did not surpass the performance of HybridMix, such as in the case of PreActResNet-50. Notably, our proposed approach achieved an absolute 2\% improvement in error rate for the CIFAR-100 dataset when evaluated using the PreActResNet-101 architecture.

\begin{table*}
\caption{Test accuracy (\%) on CIFAR10 dataset  using various models architectures.}
\begin{tabular}{l|ccc}

\hline
\multicolumn{4}{c}{\textbf{CIFAR10 Dataset}} \\ \hline 
\hline Model & ResNet-18 & ResNet-110 & Wide ResNet-28-10 \\
\hline Cutout~\cite{devries2017improved} & $95.6 \pm 0.1$ & $94.8 \pm 0.1$ & $96.9 \pm 0.1$ \\
KeepCutout~\cite{gong2021keepaugment,devries2017improved} & $96.1 \pm 0.1$ & $9 5 . 5 \pm 0 . 1$ & $\mathbf{9 7 . 3} \pm \mathbf{0 . 1}$ \\
KeepCutout (LR)~\cite{gong2021keepaugment,devries2017improved} & $\mathbf{9 6 . 2} \pm \mathbf{0 . 1}$ & $\mathbf{9 5 . 5} \pm \mathbf{0 . 1}$ & $\mathbf{9 7 . 3} \pm \mathbf{0 . 1}$ \\
KeepCutout (RL)~\cite{gong2021keepaugment,devries2017improved} & $96.0 \pm 0.1$ & $95.3 \pm 0.1$ & $97.2 \pm 0.1$ \\
\hline 
\textbf{Ours-KeepOriginalCutout } & $\mathbf{96.6 \pm 0.1}$ & $95.1 \pm 0.2$ & $97.1 \pm 0.2$ \\

 \hline Model & WideResNet-28-10 &  PyramidNet \\
\hline AutoAugment~\cite{cubuk2019autoaugment,gong2021keepaugment} & $97.3 \pm 0.1$  & 98.5 \\
KeepAutoAugment~\cite{gong2021keepaugment} & $9 7 . 8 \pm 0 . 1$  & $\mathbf{9 8 . 7} \pm \mathbf{0 . 0}$ \\
KeepAutoAugment (LR)~\cite{gong2021keepaugment} & $9 7 . 8 \pm 0 . 1$  & $\mathbf{9 8 . 7} \pm \mathbf{0 . 0}$ \\
KeepAutoAugment (EL)~\cite{gong2021keepaugment} & $9 7 . 8 \pm 0 . 1$  & $98.6 \pm 0.0$ \\ \hline 
\textbf{KeepOriginalAugment (Ours)} & $\mathbf{97.9 \pm 0.3}$  & $98.6 \pm 0.1$ \\
\hline
\end{tabular} 

\label{tab:cifar10_results}
\end{table*}

\begin{table*}
\caption{Test Error rate (\%) on different datasets  using various models architectures.}
\begin{tabular}{l|ccc}

\hline
\multicolumn{4}{c}{\textbf{CIFAR10 Dataset}} \\
Model & PreActResNet-18 & PreActResNet-50 & PreActResNet-101 \\
\hline Baseline & $5.17 \pm 0.27$ & $4.6 \pm 0.2$ & $4.49 \pm 0.18$ \\
+Cutout ~\cite{devries2017improved,choi2021salfmix} & $4.3 \pm 0.09$ & $3.77 \pm 0.08$ & $3.54 \pm 0.11$ \\
+SalfMix~\cite{choi2021salfmix} & $4 . 1 4 \pm 0 . 2 5$ & $3 . 6 1 \pm 0 . 0 9$ & $3 . 3 8 \pm 0 . 1 1$ \\
 +Mixup ~\cite{choi2021salfmix,pmlr-v97-verma19a} & $4.1 \pm 0.39$ & $3.56 \pm 0.04$ & $3.54 \pm 0.08$ \\
+SaliencyMix~\cite{uddinsaliencymix,choi2021salfmix}  & $3.8 \pm 0.1$ & $2.98 \pm 0.1$ & $2.82 \pm 0.08$ \\
+CutMix ~\cite{yun2019cutmix,choi2021salfmix} & $3.96 \pm 0.21$ & $3.07 \pm 0.09$ & $2.95 \pm 0.06$ \\
+ResizeMix ~\cite{qin2020resizemix,choi2021salfmix} & $3.74 \pm 0.2$ & $3.09 \pm 0.11$ & $2.85 \pm 0.09$ \\
+HybridMix v1~\cite{choi2021salfmix} & $3.85 \pm 0.13$ & $3.22 \pm 0.04$ & $3.04 \pm 0.14$ \\
+HybridMix v2~\cite{choi2021salfmix} & $3.74 \pm 0.05$ & $2.94 \pm 0.09$ & $2.78 \pm 0.04$ \\
+HybridMix v3~\cite{choi2021salfmix} & $3 . 3 8 \pm 0 . 0 7$ & $\mathbf{2 . 8 9 \pm 0 . 1 1}$ & $2 . 7 5 \pm 0 . 0 7$ \\
\hline 
\textbf{+Ours- KeepOriginalAugment} & $\mathbf{3.30 \pm 0.10}$ & $3.20 \pm 0.30$ & $\mathbf{2.70 \pm 0.00}$ \\ 
\hline
\multicolumn{4}{c}{\textbf{CIFAR100 Dataset}} \\ \hline 

Model & PreActResNet-18 & PreActResNet-50 & PreActResNet-101 \\
\hline Baseline & $24.22 \pm 0.22$ & $22.02 \pm 0.18$ & $21.81 \pm 0.24$ \\
+Cutout ~\cite{devries2017improved,choi2021salfmix}   & $23.72 \pm 0.27$ & $21.64 \pm 0.43$ & $21.46 \pm 0.25$ \\
+SalfMix~\cite{choi2021salfmix} & $2 2 . 6 4 \pm 0 . 1 3$ & $2 0 . 4 8 \pm 0 . 1 7$ & $1 9 . 8 9 \pm 0 . 1 3$ \\
  +Mixup ~\cite{choi2021salfmix,pmlr-v97-verma19a} & $21.78 \pm 0.4$ & $18.91 \pm 0.26$ & $18.82 \pm 0.37$ \\
+SaliencyMix~\cite{uddinsaliencymix,choi2021salfmix}  & $20.02 \pm 0.13$ & $17.5 \pm 0.16$ & $17.33 \pm 0.09$ \\
+CutMix ~\cite{yun2019cutmix,choi2021salfmix}  & $20.51 \pm 0.17$ & $17.72 \pm 0.17$ & $17.61 \pm 0.25$ \\
+ResizeMix ~\cite{qin2020resizemix,choi2021salfmix}  & $20.96 \pm 0.11$ & $17.56 \pm 0.09$ & $17.36 \pm 0.19$ \\ 

+HybridMix v1~\cite{choi2021salfmix} & $21.42 \pm 0.17$ & $18.27 \pm 0.12$ & $17.45 \pm 0.12$ \\
+HybridMix v2~\cite{choi2021salfmix} & $19.88 \pm 0.27$ & $17.38 \pm 0.27$ & $1 7 . 2 2 \pm 0 . 2 1$ \\
+HybridMix v3~\cite{choi2021salfmix} & ${1 9 . 8 4} \pm {0 . 0 9}$ & $\mathbf{1 7 . 3 \pm 0 . 2 5}$ & $17.25 \pm 0.23$ \\
\hline 
\textbf{+Ours- KeepOriginalAugment} & $\mathbf{18.70 \pm 0.30}$ & $17.80 \pm 0.2$ & $\mathbf{15.90 \pm 0.11}$ \\ 

\hline


\multicolumn{4}{c}{\textbf{TinyImageNet Dataset}} \\
\hline Model & PreActResNet-18 & PreActResNet-50 \\
\hline Baseline & $42.33 \pm 0.21$ & $38.58 \pm 0.24$ &  \\
+Cutout ~\cite{devries2017improved,choi2021salfmix} & $42.04 \pm 0.31$ & $38.36 \pm 0.21$ &  \\
+SalfMix~\cite{choi2021salfmix} & $4 0 . 2 8 \pm 0 . 2 8$ & $3 5 . 9 2 \pm 0 . 0 7$ &  \\
\hline +Mixup  ~\cite{choi2021salfmix,pmlr-v97-verma19a} & $40.22 \pm 0.2$ & $35.51 \pm 0.15$ &  \\
+SaliencyMix ~\cite{uddinsaliencymix,choi2021salfmix} & $37.76 \pm 0.05$ & $\mathbf{32.83 \pm 0.47}$ &  \\
+CutMix~\cite{yun2019cutmix,choi2021salfmix} & $38.11 \pm 0.32$ & $33.54 \pm 0.19$ &  \\
+ResizeMix ~\cite{qin2020resizemix,choi2021salfmix} & $38.47 \pm 0.25$ & $33.25 \pm 0.12$ & \\ \hline 
\textbf{+Ours- KeepOriginalAugment} & $\mathbf{35.1 \pm 0.20}$ & $35.6 \pm 0.12$  & \\ 
\hline 
\end{tabular}

\label{tab:all_datasets_results}
\end{table*}

\subsection{Discussion}\label{discussion}

In this study, we aimed to answer two important questions while addressing the limitations of SalfMix and KeepAugment. The first question was \textbf{Where to place the salient region?} Experimental results indicated that the salient region should be placed randomly in any non-salient region. The random placement of the salient region provides scaling data augmentation to that specific region. For example, on epoch 1, it may be placed in the region with the minimum area, while on the next epoch, it may be placed in a region with a different area. This random placement strategy ensures that scaling data augmentation is applied to various regions, promoting diversity in the training process.

The second question we addressed was \textbf{Which parts should be augmented?} Based on our experiments, both the salient region and the whole image should be augmented. This approach allows the neural network to access both the original data and diverse augmented data during training. By providing only augmented examples to the model, there is a concern that the model may not have access to the original data, which can lead to a shift in the data distribution. To mitigate this, we adopted a well-known technique of finding a balance between the augmented and original samples. We used a probability of 0.5 as a trade-off, meaning that during training, the neural network was fed with original and augmented data in equal proportions as it is investigated by random erasing~\cite{zhong2020random}. This dynamic combination of original and augmented data ensures that the model learns from both sources, enhancing diversity and fidelity in the learned features.
\section{Conclusion}\label{conclusion} We have introduced KeepOriginalAugment, a novel and effective method designed to address the limitations of existing state-of-the-art (SOTA) data augmentation techniques. By tackling the challenges of feature redundancy and domain shift, our method offers a promising solution for enhancing the performance and generalization of deep learning models. Through comprehensive experiments conducted on diverse datasets and utilizing various network architectures, we have demonstrated the superior performance of KeepOriginalAugment. Our method consistently outperformed previous SOTA methods by a significant margin, showcasing its ability to achieve higher accuracy, lower error rates in different datasets.  The key strength of KeepOriginalAugment lies in its optimized strategies for placing salient regions and selecting augmented parts. By providing scaling data augmentation to randomly chosen non-salient regions, we ensure the diversity and richness of training data. Additionally, by augmenting both the salient region and the entire image, we strike a balance between original and augmented samples, enabling the model to access a combination of the two and preventing potential data distribution shifts. In future, we plan to check diversity of dataset from gender and professional perspectives which can help in debiasing. 


\section{Acknowledgment}
This research was supported by Science Foundation Ireland under grant numbers 18/CRT/6223 (SFI Centre for Research Training in Artificial intelligence), SFI/12/RC/2289/$P\_2$ (Insight SFI Research Centre for Data Analytics), 13/RC/2094/$P\_2$ (Lero SFI Centre for Software) and 13/RC/2106/$P\_2$ (ADAPT SFI Research Centre for AI-Driven Digital Content Technology). For the purpose of Open Access, the author has applied a CC BY public copyright licence to any Author Accepted Manuscript version arising from this submission.

%
%
%
%

\end{document}